\documentclass{article}

\usepackage[nonatbib,final]{neurips_2023}

\usepackage[utf8]{inputenc} 
\usepackage[T1]{fontenc}    
\usepackage{hyperref}       
\usepackage{url}            
\usepackage{booktabs}       
\usepackage{multirow}
\usepackage{amsfonts}       
\usepackage{nicefrac}       
\usepackage{microtype}      
\usepackage{xcolor}         
\usepackage[draft]{fixme}
\usepackage{graphicx}
\usepackage[natbib=true, style=numeric]{biblatex}
\usepackage{cancel}
\usepackage{textgreek}
\usepackage{amsmath,amsfonts,bm}

\NewDocumentCommand{\evalat}{sO{\big}mm}{%
  \IfBooleanTF{#1}
   {\mleft. #3 \mright|_{#4}}
   {#3#2|_{#4}}%
}

\def\figref#1{figure~\ref{#1}}

\def\eqref#1{equation~\ref{#1}}

\def\1{\bm{1}}

\def\pr{\vec{r}}

\newcommand{\Ex}[1]{\left\langle{#1}\right\rangle}

\def\reta{{\textnormal{$\eta$}}}

\def\repsilon{{\textnormal{$\epsilon$}}}

\def\rvepsilon{{\mathbf{\epsilon}}}

\def\vzero{{\bm{0}}}

\def\vtheta{{\bm{\theta}}}

\def\vr{{\bm{r}}}

\def\vz{{\bm{z}}}
\def\vzero{{\bm{0}}}

\def\evr{{r}}

\DeclareMathAlphabet{\mathsfit}{\encodingdefault}{\sfdefault}{m}{sl}
\SetMathAlphabet{\mathsfit}{bold}{\encodingdefault}{\sfdefault}{bx}{n}

\def\sN{{\mathbb{N}}}

\def\sR{{\mathbb{R}}}

\newcommand{\E}{\mathbb{E}}

\newcommand{\Var}{\mathrm{Var}}

\title{Coherent energy and force uncertainty in deep learning force fields}

\author{%
  Peter Bjørn Jørgensen\thanks{Majority of the work was done while still at Technical University of Denmark.}\\
  Alexandra Institute\\
  2300 Copenhagen, Denmark \\
  \texttt{peterbjorgensen@gmail.com} \\
  \And
  Jonas Busk \\
  Technical University of Denmark \\
  2800 Kongens Lyngby \\
  \AND
  Ole Winther \\
  Technical University of Denmark \\
  2800 Kongens Lyngby \
  \And
  Mikkel N. Schmidt \\
  Technical University of Denmark \\
  2800 Kongens Lyngby \\
}

\begin{document}

\maketitle

\begin{abstract}
	In machine learning energy potentials for atomic systems, forces are commonly obtained as the negative derivative of the energy function with respect to atomic positions. To quantify aleatoric uncertainty in the predicted energies, a widely used modeling approach involves predicting both a mean and variance for each energy value.
	However, this model is not differentiable under the usual white noise assumption, so energy uncertainty does not naturally translate to force uncertainty.
	In this work we propose a machine learning potential energy model in which energy and force aleatoric uncertainty are linked through a spatially correlated noise process.
    We demonstrate our approach on an equivariant messages passing neural network potential trained on energies and forces on two out-of-equilibrium molecular datasets.
	Furthermore, we also show how to obtain epistemic uncertainties in this setting based on a Bayesian interpretation of deep ensemble models.
\end{abstract}

\section{Introduction}
In recent years, the use of machine learning force fields have enabled studies of atomic systems that are otherwise out of reach because of the computational complexity of traditional electronic structure methods such as density functional theory (DFT). Deep learning force fields are typically trained on large datasets of molecular energies and forces computed using DFT, and to increase data efficiency training data can be collected sequentially, guided by the model uncertainty. Thus, robust uncertainty estimates are crucial to optimize data collection and generally increase interpretability of predictions.

In machine learning force fields, forces are often obtained by computing the partial derivatives of the potential energy surface with respect to the atom positions, which ensures that the derived force field is conservative. Futhermore, when the potential energy model is invariant to rotations and translations, the force field is equivariant to rotations.

Mean-variance networks and deep ensemble models \citep{deepensembles} are commonly used to obtain uncertainty estimates from deep learning models \cite{carreteDeepEnsemblesVs2023, buskGraphNeuralNetwork2023}.
However, given a mean-variance network for the energy, deriving the force uncertainty through the derivative is not possible under the usual assumption of white/uncorrelated noise.
This difficulty has been noted by several authors \citep{gasteigerFastUncertaintyAwareDirectional2022, carreteDeepEnsemblesVs2023}.
\citet{gasteigerFastUncertaintyAwareDirectional2022} writes: {\em``There is thus no general way of estimating $\sigma_F$ for these kinds of models. Instead, we have to rely on $\sigma_E$ as the uncertainty measure and hope that it correlates with the force error.''} Notice that the force standard deviation $\sigma_F$ and the energy standard deviation $\sigma_E$ have different physical units, so while hoping that the two might be correlated, we would at least have to correct the units to obtain meaningful uncertainties. Concurrently with our work, \citet{carreteDeepEnsemblesVs2023} found that the force variance can not be derived directly from the mean-variance network for the energy because we need a differentiable function for the covariance of the energy observations at two arbitrary points at close proximity. They therefore workaround the problem by expanding the model to predict a separate $\sigma_F$ for each individual atom as also done in \citep{buskGraphNeuralNetwork2023}.

In this work we propose to relax the white noise assumption for the noise and derive closed form expressions for the mean and variance of the forces directly from the potential energy model. We show that a parameter naturally arises, which can be trained or adjusted post-hoc and can be understood as the squared inverse length scale of the noise process.

\section{White noise model}
Given a sequence of atomic numbers $\vz=\left( z_1, \ldots,z_i,\ldots, z_N \right) \in \sN$ and corresponding atomic positions $\vr=\left( \pr_1, \ldots \pr_i, \ldots, \pr_N \right) \in \sR^D$, the underlying assumption of the machine learning potential energy model is that the energy observations are obtained as:
\begin{align}
	E_\text{obs}(\vz,\vr) = E_\vtheta(\vz,\vr) +  \rho_\vtheta(\vz, \vr) \repsilon \, ,
	\label{eq:e_obs}
\end{align}
where $E_\vtheta(\cdot)$ and $\rho_\vtheta^2(\cdot)$ are the mean and variance outputs, respectively, of a machine learning potential with parameters $\vtheta$ and $\rvepsilon$ is zero mean, random noise of unit variance. 
The model is roto-translationally invariant when
\begin{align}
    E_\vtheta(\vz,\vr) = E_\vtheta(P\vz,RP\vr+t)
\end{align}
for any permutation $P$, rotation $R$ and translation $t$.
The typical assumption is that $\rvepsilon$ is white noise, i.e. its autocorrelation function $\E\left[ \rvepsilon(\vz, \vr) \rvepsilon(\vz', \vr') \right] $ is 1 when $(\vz, \vr)$ and $(\vz', \vr')$ represent the same configuration, $(\vz, \vr)=(P\vz', RP\vr'+t)$, and 0 otherwise.

We say that a function $f(t)$ is differentiable at a point $t=t_0$ if the limit exists:
\begin{align}
	f'(t_0) = \lim_{\Delta t \to 0} \frac{f(t_0 + \Delta t) - f(t_0)}{\Delta t}.
	\label{eq:derivative}
\end{align}
To extend the notion of a derivative to a stochastic process, we can say that a random process $X(t)$ is differentiable in the mean-squared sense at a point $t=t_0$ if there exists a stochastic process $X'(t)$ such that
\begin{align}
	\lim_{\Delta t \to 0} \E \left[ \left( \frac{X(t_0+\Delta t) - X(t_0)}{\Delta t} - X'(t_0) \right)^{2} \right]=0.
	\label{eq:mean_squared_diff}
\end{align}

If we attempt to take the partial derivative of \eqref{eq:e_obs} with respect to one of the atomic coordinates, we will see that the limit in \eqref{eq:mean_squared_diff} does not exist because of the discontinuity of the autocorrelation function of $\rvepsilon(\vz, \vr)$ with respect to $\vr$.
A typically used workaround is to model the force uncertainty independently from the energy uncertainty, i.e. we assume that the force observations
are obtained by a separate noise process, such that the observed force on the $i$th atom in the $d$th spatial dimension  is coming from this process:
\begin{align}
	f_{\text{obs}, i, d}(\vz,\vr) = - \frac{\partial E_\vtheta}{\partial \evr_{i,d}}(\vz,\vr) + \omega_{\vtheta,i}(\vz, \vr) \rvepsilon_{i,d} \, ,
	\label{eq:force_obs}
\end{align}
where $\omega^2_{\vtheta,i}$ is the variance output from the deep learning potential for each atom and $\rvepsilon_{i,d}$ is zero mean, random noise of unit variance. A simplifying assumption is that the noise level is the same in all the spatial dimensions, but the model could be extended to different noise levels in each direction using equivariant vectorial outputs from the model.

\section{Colored noise model}\label{sec:colored_model}
The form of the energy predictions of this model is the same as in \eqref{eq:e_obs} and we use the same symbols for the outputs of the machine learning potential, even though they are not directly transferable from one setting to the other. The energy observations are thus obtained as:
\begin{align}
	E_\text{obs}(\vz,\vr) = E_\vtheta(\vz,\vr) +  \rho_\vtheta (\vz, \vr) \reta \, ,
	\label{eq:e_obs_colored}
\end{align}
where $\reta$ is a stochastic process with a differentiable autocorrelation function $\E_{\reta \reta}\left[ \reta(\vz,\vr) \reta(\vz',\vr')  \right]$ function with respect to the atom positions $\vr$ and $\vr'$ and $\reta$ is zero mean and unit variance for all $\vz$ and $\vr$.
The mean and variance for $E_\text{obs}(\vz,\vr)$ are still given directly from the machine learning potential outputs $E_\vtheta(\vz,\vr)$ and $\rho_\vtheta^2 (\vz, \vr)$, respectively.
Another simplifying assumption is that the noise process $\reta$ is wide-sense stationary, i.e. its mean does not change with $\vz,\vr$ and the autocorrelation function is (locally) a function of the difference between the atom positions  $\E_{\reta \reta}\left[ \reta(\vz,\vr) \reta(\vz',\vr')  \right]=R_{\reta \reta}^{\vz\vz'}(\vr-\vr')$.
The force mean is the same as in \eqref{eq:force_obs}, namely $- \frac{\partial E_\vtheta}{\partial \evr_{i,d}}(\vz,\vr)$, while we get the following expression for the force variance (derived in appendix \ref{app:cov_energy_force}):
\begin{align}
	\Var\left(- \frac{\partial E_{\text{obs}}}{\partial \evr_{i,d}}(\vz,\vr)\right) = - \evalat[\Big]{\frac{\partial^2 R_{\reta \reta}^{\vz \vz'}(\Delta \vr)}{\partial \Delta r_{i,d}^2}}{\Delta\vr=\vzero} \rho_\vtheta^2(\vz,\vr) + R_{\reta \reta}^{\vz \vz'}(\vzero) \left(  \frac{\partial \rho_\vtheta(\vz,\vr)}{\partial r_{i,d}}\right)^2 \, .
	\label{eq:force_var_colored}
\end{align}
Notice that the force variance only depends on the noise autocorrelation through its value at $\Delta\vr=0$ and its second derivative at $\Delta\vr=0$. Since $\reta$ is zero mean and unit variance we get $R_{\reta \reta}^{\vz \vz'}(\vzero)=1$.
If $\eta$ were a Gaussian process with the exponentiated quadratic kernel $R(\vr,\vr')=\exp \left( \frac{||\vr-\vr'||^2}{2 \ell^2} \right)$ we would refer to $\gamma=- \evalat[\Big]{\frac{\partial^2 R_{\reta \reta}^{\vz \vz'}(\Delta \vr)}{\partial \Delta r_{i,d}^2}}{\Delta\vr=\vzero}=\frac{1}{\ell^2}$ as the inverse {\em length scale} squared of the kernel. Hence, when we train our neural network potential we can treat the inverse length scale as a hyperparameter $\hat{\gamma}$ and estimate it from the training data. The force variance thus becomes:
\begin{align}
	\Var\left(- \frac{\partial E_{\text{obs}}}{\partial \evr_{i,d}}(\vz,\vr)\right) = \hat{\gamma} \rho_\vtheta^2(\vz,\vr) + \left(  \frac{\partial \rho_\vtheta(\vz,\vr)}{\partial r_{i,d}}\right)^2 \, .
	\label{eq:force_var_colored_simplified}
\end{align}
If we drop the wide-sense stationary assumption the model can be generalised to multiple inverse length scales and we would predict the $\gamma$ parameters as another output of the model for each atom, i.e. $\hat{\gamma}_{\vtheta, i}(\vz,\vr)$, but we would also need to predict another quantity of the noise process, specifically $\zeta = \evalat[\Big]{\frac{\partial R_{\reta \reta}^{\vz \vz'}(\vr,\vr')}{\partial r'_{i,d}}}{\vr'=\vr}$ (see \eqref{eq:force_var_full} in appendix \ref{app:cov_energy_force}).
To get epistemic uncertainty predictions from the model we use a Bayesian interpretation of deep ensemble models~\citep{wilsonBayesianDeepLearning2020, hoffmannDeepEnsemblesBayesian2021, gustaffsonScalableDeepLearning}. See \autoref{app:epistemic_uncertainty}.

\section{Toy Example}
\begin{figure}[tbp]
	\centering\includegraphics[width=1.0\textwidth]{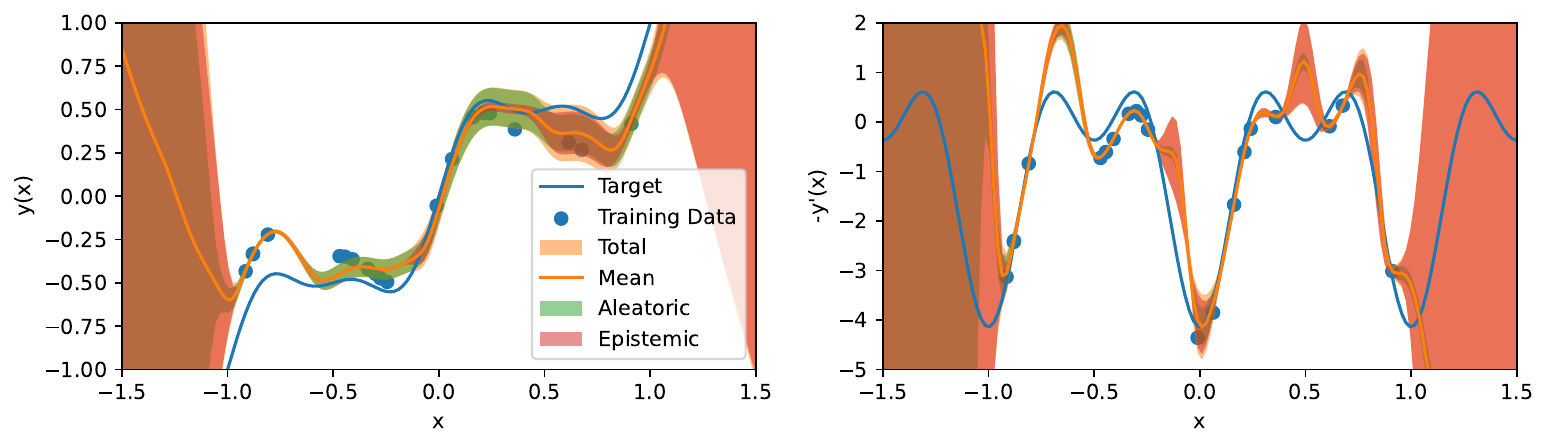}
	\caption{Ensemble model (with 10 instances) fitted to energy (left) and force (right) observations with correlated noise.}
	\label{fig:toy_example}
\end{figure}

In this example we fit an ensemble of multilayer perceptron neural networks to a synthetic 1-dimensional dataset that is generated using the same underlying assumptions as the model.
The noise free energy function is $y(x)= x + 0.3 \sin(2 \pi x) + 0.1 \sin(4\pi x)$.
The additive noise is sampled from a Gaussian process with kernel $k(a,b)=\alpha^2 \left(\exp \left( -\frac{(a-b)^2}{2 \ell^2} \right) + \delta(a-b)10^{-4}\right)$ with $\alpha=0.2$ and $\ell=0.5$. We sample 20 training examples uniformly between $-1$ and $1$ and the resulting predictions are shown in \figref{fig:toy_example}.
The individual models are trained with maximum likelihood using a normal distribution parameterised with the mean and variance expressions derived in section \ref{sec:colored_model}.
As expected, the predicted mean function does not follow the true energy function because the added noise function is a single realisation from a Gaussian process prior.
The aleatoric uncertainty is small when the samples are widely spaced, but increases when the samples are close together, which allows the model to trade precision for smoothness.
We also see that the epistemic uncertainty is large in regions where there are no training samples and decreases as we get closer to the training samples, while the aleatoric uncertainty extrapolates poorly outside of the training interval
with high uncertainty for negative $x$ and small uncertainty for positive $x$.

\section{Application to Molecular Data}
We apply the proposed method to an ensemble of equivariant message passing neural networks, specifically an extension of the PaiNN~\citep{painn} model.
The proposed method is evaluated on two publicly available datasets designed specifically for the development and evaluation of ML potentials, ANI-1x~\citep{ani1x} and Transitions1x~\citep{schreiner2022}.
The datasets include different compositions and out-of-equilibrium structures and contain a wide distribution of energies and forces. See \autoref{app:datasets} for details.

For comparison we also train a \emph{vanilla} ensemble model, consisting of 5 models without variance outputs, but it still has epistemic uncertainty given by the variances across the ensemble.
We also compare with a \emph{white noise} ensemble model, that uses the usual assumption of uncorrelated noise.
The results are summarised in Table~\ref{tab:results}. The symbol $\downarrow$ means that lower values are better. The metrics are described in detail in \autoref{app:reliability}.
\begin{table*}
\footnotesize
\caption{Test results of ensemble models ($M=5$) with different noise assumptions trained on the ANI-1x (A1x) and Transition1x (T1x) datasets.
Energy errors are averaged over molecules, while force errors are computed component-wise and averaged over the spatial dimensions and atoms.}
\label{tab:results}
\begin{tabular*}{\textwidth}{@{\extracolsep{\fill}}rlrrrrrrr}
\toprule
	\multicolumn{1}{c}{\bf Data} & {\bf Model} & \multicolumn{7}{c}{\bf Energy (eV)} \\ 
\cmidrule(lr){3-9}
	&  & MAE$\downarrow$ & RMSE$\downarrow$ & NLL$\downarrow$ & RZV & ENCE$\downarrow$ & CV & RMV \\
\midrule
\multirow{3}*{A1x} & Vanilla &
0.013 & 0.031 & -2.13 & 1.83 & 0.77 & 1.05 & 0.016\\
& White &
0.010 & 0.028 & -3.13 & 0.69 & 0.31  & 1.54 & 0.033 \\
& Colored &
0.012 & 0.026 & -2.56 & 0.42 & 0.56  & 0.97 & 0.048 \\
\midrule
\multirow{3}*{T1x} & Vanilla &
0.033 & 0.060 & -0.83 & 2.10 &  1.06 & 0.73 & 0.029 \\
& White &
0.039 & 0.064 & -1.67 & 1.04 &  0.18 & 0.47 & 0.051 \\
& Colored &
0.037 & 0.063 & -1.75 & 0.93 &  0.21 & 0.48 & 0.054 \\
\midrule
	&  & \multicolumn{7}{c}{\bf Forces (eV)} \\ 
\cmidrule(lr){3-9}
&  & MAE$\downarrow$ & RMSE$\downarrow$ & NLL$\downarrow$ & RZV & ENCE$\downarrow$ & CV & RMV \\
\midrule
\multirow{3}*{A1x} & Vanilla &
0.018 & 0.037 & -1.74 & 1.89 & 0.76 & 1.31  & 0.027 \\
& White &
0.017 & 0.041 & -2.67 & 0.70 & 0.29  & 1.32 & 0.052 \\
& Colored &
0.018 & 0.039 & -2.51 & 0.66 & 0.32 & 1.13 & 0.051 \\
\midrule
\multirow{3}*{T1x} & Vanilla &
0.037 & 0.075 & -1.56 & 1.54 &  0.47 & 1.08 & 0.057 \\
& White &
0.038 & 0.076 & -1.86 & 0.86 &  0.18 & 0.84 & 0.076 \\
& Colored &
0.038 & 0.075 & -1.78 & 0.79 & 0.24 & 0.76 & 0.074 \\
\bottomrule
\end{tabular*}
\end{table*}
In general we see that the accuracy in terms of mean absolute error (MAE) and root mean squared error (RMSE) of all three model variants are in the same ballpark and comparable to previous work~\citep{buskGraphNeuralNetwork2023}, i.e. we do not need to sacrifice prediction accuracy to include aleatoric uncertainty in the model ensemble.
On average the vanilla ensemble underestimates the errors (RZV>1) while the other two models overestimate the errors (except white noise model in T1x energy) and have lower (better) negative log-likelihood.
The white noise model is better calibrated (lower ENCE and NLL) than the colored noise model. The coefficient of variation (CV) is lower for the colored noise model. Deriving the energy and force uncertainty from a single output perhaps makes the model less expressive. However, the reliability diagrams \autoref{app:reliability} shows that the uncertainties are well-behaved and it should be possible to improve the uncertainty estimates by calibrating the uncertainties on the validation set as in \cite{buskGraphNeuralNetwork2023}.


\section{Conclusion}
We have presented a method for training single and ensemble models with coherent uncertainty estimates for energy and forces of atomic systems.
This allows coherent training of energy and force uncertainty from a single output of the model. However, a parameter related to the length scale of the noise arises, which needs to be tuned or learned.
In this way, our method is akin to Gaussian process regression, but importantly it does not require the design of a kernel function and is therefore easier to apply with existing state of the art deep learning potentials.
The method is not limited to deep learning force fields, but can be valuable in other domains where both the energy and gradients are observed.
We hope that this perspective on uncertainty quantification will contribute to further development of machine learning force fields with theoretically grounded energy and force uncertainties.

\printbibliography

@article{pernot2022,
author = {Pernot,Pascal },
title = {Prediction uncertainty validation for computational chemists},
journal = {The Journal of Chemical Physics},
volume = {157},
number = {14},
pages = {144103},
year = {2022},
doi = {10.1063/5.0109572},
URL = {https://doi.org/10.1063/5.0109572},
eprint = {https://doi.org/10.1063/5.0109572}
}

@misc{gasteigerFastUncertaintyAwareDirectional2022,
  title = {Fast and {{Uncertainty-Aware Directional Message Passing}} for {{Non-Equilibrium Molecules}}},
  author = {Gasteiger, Johannes and Giri, Shankari and Margraf, Johannes T. and G{\"u}nnemann, Stephan},
  year = {2022},
  month = apr,
  number = {arXiv:2011.14115},
  eprint = {arXiv:2011.14115},
  publisher = {{arXiv}},
  urldate = {2023-03-23},
  archiveprefix = {arxiv}
}

@article{busk2021,
doi = {10.1088/2632-2153/ac3eb3},
url = {https://dx.doi.org/10.1088/2632-2153/ac3eb3},
year = {2021},
month = {dec},
publisher = {IOP Publishing},
volume = {3},
number = {1},
pages = {015012},
author = {Jonas Busk and Peter Bjørn Jørgensen and Arghya Bhowmik and Mikkel N Schmidt and Ole Winther and Tejs Vegge},
title = {Calibrated uncertainty for molecular property prediction using ensembles of message passing neural networks},
journal = {Machine Learning: Science and Technology},
}

@inproceedings{seitzer2022betanll,
title={On the Pitfalls of Heteroscedastic Uncertainty Estimation with Probabilistic Neural Networks},
author={Maximilian Seitzer and Arash Tavakoli and Dimitrije Antic and Georg Martius},
booktitle={International Conference on Learning Representations},
year={2022},
url={https://openreview.net/forum?id=aPOpXlnV1T}
}

@article{deepensembles,
  title={Simple and scalable predictive uncertainty estimation using deep ensembles},
  author={Lakshminarayanan, Balaji and Pritzel, Alexander and Blundell, Charles},
  journal={Advances in neural information processing systems},
  volume={30},
  pages={6402--6413},
  year={2017}
}

@inproceedings{painn,
  title={Equivariant message passing for the prediction of tensorial properties and molecular spectra},
  author={Sch{\"u}tt, Kristof and Unke, Oliver and Gastegger, Michael},
  booktitle={International Conference on Machine Learning},
  pages={9377--9388},
  year={2021},
  organization={PMLR}
}

@article{ani1x,
author = {Smith,Justin S.  and Nebgen,Ben  and Lubbers,Nicholas  and Isayev,Olexandr  and Roitberg,Adrian E. },
title = {Less is more: Sampling chemical space with active learning},
journal = {The Journal of Chemical Physics},
volume = {148},
number = {24},
pages = {241733},
year = {2018},
doi = {10.1063/1.5023802},
eprint = {https://doi.org/10.1063/1.5023802}
}

@Article{schreiner2022,
author={Schreiner, Mathias
and Bhowmik, Arghya
and Vegge, Tejs
and Busk, Jonas
and Winther, Ole},
title={Transition1x - a dataset for building generalizable reactive machine learning potentials},
journal={Scientific Data},
year={2022},
month=dec,
day={24},
volume={9},
number={1},
pages={779},
issn={2052-4463},
doi={10.1038/s41597-022-01870-w},
url={https://doi.org/10.1038/s41597-022-01870-w}
}

@Article{levi2022,
AUTHOR = {Levi, Dan and Gispan, Liran and Giladi, Niv and Fetaya, Ethan},
TITLE = {Evaluating and Calibrating Uncertainty Prediction in Regression Tasks},
JOURNAL = {Sensors},
VOLUME = {22},
YEAR = {2022},
NUMBER = {15},
ARTICLE-NUMBER = {5540},
PubMedID = {35898047},
ISSN = {1424-8220},
DOI = {10.3390/s22155540}
}

@inproceedings{wilsonBayesianDeepLearning2020,
  title = {Bayesian Deep Learning and a Probabilistic Perspective of Generalization},
  booktitle = {Proceedings of the 34th {{International Conference}} on {{Neural Information Processing Systems}}},
  author = {Wilson, Andrew Gordon and Izmailov, Pavel},
  year = {2020},
  month = dec,
  series = {{{NIPS}}'20},
  pages = {4697--4708},
  publisher = {{Curran Associates Inc.}},
  address = {{Red Hook, NY, USA}},
  isbn = {978-1-71382-954-6}
}

@misc{hoffmannDeepEnsemblesBayesian2021,
  title = {Deep {{Ensembles}} from a {{Bayesian Perspective}}},
  author = {Hoffmann, Lara and Elster, Clemens},
  year = {2021},
  month = nov,
  number = {arXiv:2105.13283},
  eprint = {2105.13283},
  eprinttype = {arxiv},
  primaryclass = {cs, stat},
  publisher = {{arXiv}},
  doi = {10.48550/arXiv.2105.13283},
  archiveprefix = {arXiv}
}

@INPROCEEDINGS{gustaffsonScalableDeepLearning,
author = {F. K. Gustafsson and M. Danelljan and T. B. Schon},
booktitle = {2020 IEEE/CVF Conference on Computer Vision and Pattern Recognition Workshops (CVPRW)},
title = {Evaluating Scalable Bayesian Deep Learning Methods for Robust Computer Vision},
year = {2020},
volume = {},
issn = {},
pages = {1289-1298},
doi = {10.1109/CVPRW50498.2020.00167},
url = {https://doi.ieeecomputersociety.org/10.1109/CVPRW50498.2020.00167},
publisher = {IEEE Computer Society},
address = {Los Alamitos, CA, USA},
month = jun
}

@misc{carreteDeepEnsemblesVs2023,
  title = {Deep {{Ensembles}} vs. {{Committees}} for {{Uncertainty Estimation}} in {{Neural-Network Force Fields}}: {{Comparison}} and {{Application}} to {{Active Learning}}},
  shorttitle = {Deep {{Ensembles}} vs. {{Committees}} for {{Uncertainty Estimation}} in {{Neural-Network Force Fields}}},
  author = {Carrete, Jes{\'u}s and {Montes-Campos}, Hadri{\'a}n and Wanzenb{\"o}ck, Ralf and Heid, Esther and Madsen, Georg K. H.},
  year = {2023},
  month = feb,
  number = {arXiv:2302.08805},
  eprint = {2302.08805},
  eprinttype = {arxiv},
  primaryclass = {physics},
  publisher = {{arXiv}},
  doi = {10.48550/arXiv.2302.08805},
  archiveprefix = {arXiv}
}

@article{buskGraphNeuralNetwork2023,
  title = {Graph Neural Network Interatomic Potential Ensembles with Calibrated Aleatoric and Epistemic Uncertainty on Energy and Forces},
  author = {Busk, Jonas and N.~Schmidt, Mikkel and Winther, Ole and Vegge, Tejs and Bj{\o}rn~J{\o}rgensen, Peter},
  year = {2023},
  journal = {Physical Chemistry Chemical Physics},
  volume = {25},
  number = {37},
  pages = {25828--25837},
  publisher = {{Royal Society of Chemistry}},
  doi = {10.1039/D3CP02143B}
}

@misc{kingmaAdamMethodStochastic2017,
  title = {Adam: {{A Method}} for {{Stochastic Optimization}}},
  shorttitle = {Adam},
  author = {Kingma, Diederik P. and Ba, Jimmy},
  year = {2017},
  month = jan,
  number = {arXiv:1412.6980},
  eprint = {1412.6980},
  primaryclass = {cs},
  publisher = {{arXiv}},
  doi = {10.48550/arXiv.1412.6980},
  urldate = {2023-10-09},
  archiveprefix = {arxiv},
}

@inproceedings{skafteReliableTrainingEstimation2019b,
  title = {Reliable Training and Estimation of Variance Networks},
  booktitle = {Advances in {{Neural Information Processing Systems}}},
  author = {Skafte, Nicki and {J{\o}rgensen}, Martin and Hauberg, S{\o}ren},
  year = {2019},
  volume = {32},
  publisher = {{Curran Associates, Inc.}},
  urldate = {2023-10-09},
}

\appendix

\section{Covariance of energy and force observations}\label{app:cov_energy_force}
We can write the energy function in the following form:
\begin{equation}
	E(x) = \mu_{\theta}(x) + \varepsilon(x) \sigma_{\theta}(x)
	\label{eq:energy} \, ,
\end{equation}
where $\varepsilon(x)$ is a stochastic process with zero mean, unit variance and differentiable autocorrelation function $R_{\varepsilon \varepsilon}(x, x')=\E\left[ \varepsilon(x) \varepsilon(x') \right]$.
The parameters $\theta$ are random variables but not dependent on $x$.
The covariance function is:
\begin{align}
	\mathrm{Cov}(E(x), E(x')) &= \mathbb{E}\left[ (\mu_{\theta}(x) + \varepsilon(x) \sigma_{\theta}(x))(\mu_{\theta}(x') + \varepsilon(x') \sigma_{\theta}(x')) \right]\nonumber \\ & \phantom{=} - \mathbb{E}\left[ \mu_\theta(x) \right]\mathbb{E}\left[ \mu_\theta(x') \right] \\
	&= \mathbb{E}\left[ (\mu_{\theta}(x) \mu_\theta(x') + \mu_\theta(x) \varepsilon(x') \sigma_\theta(x') + \mu_\theta(x') \varepsilon(x) \sigma_\theta(x)+ \varepsilon(x) \varepsilon(x') \sigma_\theta(x) \sigma_\theta(x') \right]\nonumber \\ & \phantom{=} - \mathbb{E}\left[ \mu_\theta(x) \right]\mathbb{E}\left[ \mu_\theta(x') \right] \\
	&= \mathbb{E}\left[ (\mu_{\theta}(x) \mu_\theta(x') + \cancel{\mu_\theta(x) \varepsilon(x') \sigma_\theta(x')} + \cancelto{\mathbb{E}(\varepsilon)=0}{\mu_\theta(x') \varepsilon(x) \sigma_\theta(x)}+ \varepsilon(x) \varepsilon(x') \sigma_\theta(x) \sigma_\theta(x') \right]\nonumber \\ & \phantom{=} - \mathbb{E}\left[ \mu_\theta(x) \right]\mathbb{E}\left[ \mu_\theta(x') \right] \\
	&= R_{\varepsilon \varepsilon}(x,x') \mathbb{E}\left[ \sigma_\theta(x) \sigma_\theta(x') \right] + \mathrm{Cov}(\mu_\theta(x), \mu_\theta(x'))
	\label{eq:e_cov}
\end{align}
\begin{align}
	\mathrm{Var}(E(x)) &= \lim_{x' \to x} \mathrm{Cov}(E(x), E(x')) \\
	&= R_{\varepsilon \varepsilon}(x,x) \mathbb{E}\left[ \sigma^2_\theta(x) \right] + \mathrm{Var}(\mu_\theta(x)) \, .
	\label{eq:e_var}
\end{align}
To do the covariance of the force we use that:
\begin{align}
	\mathrm{Cov}\left(\frac{\partial E(x)}{\partial x}, \frac{\partial E(x')}{\partial x'}\right) &= \frac{\partial^2 \mathrm{Cov}(E(x), E(x'))}{\partial x \partial x'} \, .
\end{align}
First take the partial derivative with respect to $x$:
\begin{align}
	\mathrm{Cov}\left(\frac{\partial E(x)}{\partial x}, {E(x')}\right) &= \frac{\partial \mathrm{Cov}(E(x), E(x'))}{\partial x }\\
	&= \frac{\partial R_{\varepsilon \varepsilon}(x,x') \mathbb{E}\left[ \sigma_\theta(x) \sigma_\theta(x') \right]}{\partial x} + \mathrm{Cov}\left( \frac{\partial \mu_\theta(x)}{\partial x}, \mu_\theta(x') \right) \\
	&= \frac{\partial R_{\varepsilon \varepsilon}(x,x')}{\partial x} \mathbb{E}\left[ \sigma_\theta(x) \sigma_\theta(x') \right] \\
	&\phantom{=}+ R_{\varepsilon \varepsilon}(x,x') \mathbb{E}\left[ \frac{\partial\sigma_\theta(x)}{\partial x} \sigma_\theta(x') \right] \\
	&\phantom{=} + \mathrm{Cov}\left( \frac{\partial \mu_\theta(x)}{\partial x}, \mu_\theta(x') \right) \, .
\end{align}
We have used the product rule for differentiation $f(x)=u(x)v(x) \Rightarrow f'(x)=u(x)v'(x)+u'(x)v(x)$. Now also take the partial derivative with respect to $x'$:
\begin{align}
	\mathrm{Cov}\left(\frac{\partial E(x)}{\partial x}, \frac{\partial E(x')}{\partial x'}\right) &= \frac{\partial^2 \mathrm{Cov}(E(x), E(x'))}{\partial x \partial x'}\\
	&= \frac{\partial^2 R_{\varepsilon \varepsilon}(x,x')}{\partial x \partial x'} \mathbb{E}\left[ \sigma_\theta(x) \sigma_\theta(x') \right] \nonumber \\
	&\phantom{=}+ \frac{\partial R_{\varepsilon \varepsilon}(x,x')}{\partial x} \mathbb{E}\left[ \frac{\sigma_\theta(x) \partial \sigma_\theta(x')}{\partial x'}  \right] \nonumber\\
	&\phantom{=}+\frac{\partial R_{\varepsilon \varepsilon}(x,x')}{\partial x'} \mathbb{E}\left[ \frac{\partial \sigma_\theta(x)  \sigma_\theta(x')}{\partial x}  \right]\nonumber \\
	&\phantom{=}+R_{\varepsilon \varepsilon}(x,x')\mathbb{E}\left[ \frac{\partial \sigma_\theta(x)  \partial \sigma_\theta(x')}{\partial x \partial x'}  \right] \nonumber \\
	&\phantom{=} + \mathrm{Cov}\left( \frac{\partial \mu_\theta(x)}{\partial x}, \frac{\partial \mu_\theta(x')}{\partial x'} \right) \, .
 \label{eq:force_cov}
\end{align}

To compute the variance we let $x' \to x$ and evaluate the expression in \eqref{eq:force_cov}.
\begin{align}
	\mathrm{Var}\left(\frac{\partial E(x)}{\partial x}\right) &= \frac{\partial^2 \mathrm{Cov}(E(x), E(x'))}{\partial x \partial x'}\\
	&= \evalat{\frac{\partial^2 R_{\varepsilon \varepsilon}(x,x')}{\partial x \partial x'}}{x'=x} \mathbb{E}\left[ \sigma_\theta(x)^2 \right] \nonumber \\
	&\phantom{=}+ 2 \evalat{\frac{\partial R_{\varepsilon \varepsilon}(x,x')}{\partial x'}}{x'=x} \mathbb{E}\left[ \evalat{ \frac{\sigma_\theta(x) \partial \sigma_\theta(x')}{\partial x'}}{x'=x}  \right] \nonumber\\
	&\phantom{=}+R_{\varepsilon \varepsilon}(x,x)\mathbb{E}\left[ \left( \frac{\partial \sigma_\theta(x)  }{\partial x }  \right)^2 \right] \nonumber \\
	&\phantom{=} + \mathrm{Var}\left( \frac{\partial \mu_\theta(x)}{\partial x} \right) \, . \label{eq:force_var_full}
\end{align}

If we further assume that the noise process $\varepsilon (x)$ is wide sense stationary, i.e.~its autocorrelation function function is a function of the difference between the inputs $R_{\varepsilon \varepsilon}(x,x')= R_{\varepsilon \varepsilon}(x-x')$, we can simplify the expression further.
Since $R_{\varepsilon \varepsilon}(x-x')$ has a maximum at $0$ it's first partial derivatives will be $0$ and for an even function $f(x)$ we have that:
\begin{align}
	\evalat{\frac{\partial^2 f(x-x')}{\partial x \partial x'}}{x'=x} = - \evalat{\frac{\partial^2 f(x)}{\partial x^2}}{x=0}
	\label{eq:diff_even_func}
\end{align}
Under the wide sense stationary noise assumption the variance of the gradient can therefore be simplified as:
\begin{align}
	\mathrm{Var}\left( \frac{\partial E(x)}{\partial x} \right) = 
	- \evalat[\Big]{\frac{\partial^2 R_{\varepsilon \varepsilon}(x)}{\partial x^2}}{x=0} \mathbb{E}\left[ \sigma_\theta^2(x) \right]
	+ R_{\varepsilon \varepsilon}(0)\mathbb{E}\left[ \left( \frac{\partial \sigma_\theta(x)}{\partial x} \right)^2  \right]
	+ \mathrm{Var}\left( \frac{\partial \mu_\theta(x)}{\partial x} \right) \, .
	\label{eq:var}
\end{align}

\subsection{First principles proof for the variance of the gradient}
Again look at \eqref{eq:energy}. It can be shown that:
\begin{equation}
	\mathrm{Var}\left(\frac{\partial E(x)}{\partial x}\right) = \mathrm{Var}\left(\frac{\partial \mu_\theta(x)}{\partial x}\right) + \mathrm{Var}\left({\frac{\partial (\varepsilon(x) \sigma_\theta(x))}{\partial x}}\right) \, ,
\end{equation}
by showing that the covariance between the first and second term is 0.
The first term (variance of the gradient of the mean) is straightforward to compute. Here we will focus on the second term.
The average of the gradient is 0, so the variance is equal to the second moment:
\begin{align}
	\mathrm{Var}\left({\frac{\partial (\varepsilon(x) \sigma_\theta(x))}{\partial x}}\right) &= \mathbb{E}\left[ \left( {\frac{\partial (\varepsilon(x) \sigma_\theta(x))}{\partial x}} \right)^2 \right] \\
	&= \mathbb{E}\left[ \lim_{\Delta x \to 0} \left(\frac{ \sigma_\theta(x+\Delta x)\varepsilon(x+\Delta x) - \sigma_\theta(x) \varepsilon(x)}{\Delta x} \right)^2 \right]\\
	&= \mathbb{E}\left[ \lim_{\Delta x \to 0} \frac{ \sigma^2_\theta(x+\Delta x)\varepsilon^2(x+\Delta x) + \sigma^2_\theta(x) \varepsilon^2(x)- 2 \sigma_\theta(x+\Delta x)\varepsilon(x+\Delta x) \sigma_\theta(x) \varepsilon(x)}{\Delta x^2}  \right] \, .
\end{align}
Interchange the order of expectation and limit: 
\begin{align}
	&= \lim_{\Delta x \to 0} \frac{ \Ex{\sigma_\theta^2(x+\Delta x)}R_{\varepsilon \varepsilon}(0) + \Ex{\sigma^2_\theta(x)} R_{\varepsilon \varepsilon}(0) - 2 \Ex{\sigma_\theta(x+\Delta x) \sigma_\theta(x)} R_{\varepsilon \varepsilon}(\Delta x) }{\Delta x^2} \, ,  
\end{align}
where $\Ex{\cdot}=\E_{p(\theta)}\left[{\cdot}\right]$ and we have used wide sense stationary property of $\varepsilon(x)$, thus $\mathbb{E}\left[ \varepsilon^2(x) \right]= R_{\varepsilon \varepsilon}(0) \forall x \in \mathbb{R}$.
We now add $-2 \Ex{\sigma_\theta(x+\Delta x)}\sigma_\theta(x) R_{\varepsilon \varepsilon}(0)+2 \Ex{\sigma_\theta(x+\Delta x)}\sigma_\theta(x) R_{\varepsilon \varepsilon}(0)=0$ to the numerator to complete the square.

\begin{align}
	&= \lim_{\Delta x \to 0} \frac{ \Ex{(\sigma_\theta(x+\Delta x)-\sigma_\theta(x))^2} R_{\varepsilon \varepsilon}(0) + 2 \Ex{\sigma_\theta(x+ \Delta x) \sigma_\theta(x)}(R_{\varepsilon \varepsilon}(0) - R_{\varepsilon \varepsilon}(\Delta x)) }{\Delta x^2}  \\
	&= R_{\varepsilon \varepsilon}(0)\E_{p(\theta)}\left[{ \lim_{\Delta x \to 0} \left( \frac{ \sigma_\theta(x+\Delta x)-\sigma_\theta(x) }{\Delta x}\right)^2}\right] 
	+\lim_{\Delta x \to 0}\frac{2 \Ex{\sigma_\theta(x+ \Delta x) \sigma_\theta(x)}(R_{\varepsilon \varepsilon}(0) - R_{\varepsilon \varepsilon}(\Delta x)) }{\Delta x^2} \, .
\end{align}
The first term is $R_{\varepsilon \varepsilon}(0)\mathbb{E}\left[ \left( \frac{\partial \sigma_\theta(x)}{\partial x} \right)^2  \right]$. 
Let us now rearrange the second term:
\begin{align}
	&\lim_{\Delta x \to 0}\frac{2 \Ex{\sigma_\theta(x+ \Delta x) \sigma_\theta(x)}(R_{\varepsilon \varepsilon}(0) - R_{\varepsilon \varepsilon}(\Delta x)) }{\Delta x^2}\\
	&=\lim_{\Delta x \to 0}-\frac{\Ex{\sigma_\theta(x+ \Delta x) \sigma_\theta(x)}(2R_{\varepsilon \varepsilon}(\Delta x) -2R_{\varepsilon \varepsilon}(0))}{\Delta x^2}\\
	&=\lim_{\Delta x \to 0}-\frac{\Ex{\sigma_\theta(x+ \Delta x) \sigma_\theta(x)}(R_{\varepsilon \varepsilon}(-\Delta x) -2R_{\varepsilon \varepsilon}(0)+R_{\varepsilon \varepsilon}(\Delta x))}{\Delta x^2}\\
	&=- \lim_{\Delta x \to 0}\Ex{\sigma_\theta(x+ \Delta x) \sigma_\theta(x)}\lim_{\Delta x \to 0}\frac{R_{\varepsilon \varepsilon}(-\Delta x) -2R_{\varepsilon \varepsilon}(0)+R_{\varepsilon \varepsilon}(\Delta x)}{\Delta x^2}\\
	&=- \mathbb{E}\left[ \sigma^2_\theta(x) \right]\lim_{\Delta x \to 0}\frac{R_{\varepsilon \varepsilon}(-\Delta x) -2R_{\varepsilon \varepsilon}(0)+R_{\varepsilon \varepsilon}(\Delta x)}{\Delta x^2}\\
	&=- \mathbb{E}\left[ \sigma^2_\theta(x) \right] \evalat{\frac{\partial^2 R_{\varepsilon \varepsilon}(x)}{\partial x^2}}{x=0} \, .
\end{align}
Here we have used that the correlation function is even $R_{\varepsilon \varepsilon}(x)=R_{\varepsilon \varepsilon}(-x)$ and recognized the second symmetric derivative. 
These expressions are the same as found in \eqref{eq:var}.

\section{Epistemic Uncertainty}\label{app:epistemic_uncertainty}
Using a Bayesian interpretation of deep ensemble models~\citep{wilsonBayesianDeepLearning2020, hoffmannDeepEnsemblesBayesian2021, gustaffsonScalableDeepLearning} we can interpret the model weights $\vtheta^{(m)}$ of each ensemble member, $m$, as samples from an approximate posterior distribution $q(\vtheta) \approx p(\vtheta | \mathcal{D})$, where $\mathcal{D}$ is the training set.
For a regression model with input $x^*$ and output $y^*$ trained on $\mathcal{D}$ we have:
\begin{align}
	p(y^*|x^*, \mathcal{D}) &= \int p(y^*|x^*, \vtheta)p(\vtheta| \mathcal{D})d\vtheta, \\
	&\approx \frac{1}{M} \sum_{m=1}^M p(y^*|x^*, \vtheta),\quad \vtheta^{(m)} \sim p(\vtheta | \mathcal{D}) \, , \\
	&\approx \frac{1}{M} \sum_{m=1}^M p(y^*|x^*, \vtheta),\quad \vtheta^{(m)} \sim q(\vtheta) \, .
\end{align}
The first approximation is to estimate the integral with $M$ samples from the distribution $p(\vtheta | \mathcal{D})$ and the second approximation comes from approximating the true posterior $p(\vtheta | \mathcal{D})$ with the distribution $q(\vtheta)$. The uncertainty arising from $p(y^*|x^*, \vtheta)$ is the aleatoric uncertainty while the epistemic uncertainty is modeled as the uncertainty arising from the distribution of the model parameters $q(\vtheta)$.
When we train an ensemble of models using the colored noise model presented in section \ref{sec:colored_model} as the base model, we get the following expressions for the energy prediction mean and variance:
\begin{align}
	\E_\vtheta \left[ E_\text{obs}(\vz, \vr) \right] &= \E_\vtheta\left[ E_\vtheta(\vz,\vr) \right], \\
	\Var_\vtheta\left( E_\text{obs}(\vz, \vr) \right) &= \underbrace{\E_\vtheta\left[ \rho_\vtheta^2(\vz,\vr) \right]}_{\text{aleatoric}} + \underbrace{\Var_\vtheta\left( E_\vtheta(\vz,\vr)  \right)}_{\text{epistemic}}.
	\label{eq:energy_mean_var_full}
\end{align}
The expression for the variance can be derived by applying the rule of total variance $\Var(Y)=\E\left[ \Var(Y|X) \right]+\Var(\E[Y|X])$.
Similarly for the force variance we get:

\begin{align}
	\Var\left(- \frac{\partial E_{\text{obs}}}{\partial \evr_{i,d}}(\vz,\vr)\right) = %
	\underbrace{\E_\vtheta \left[ \hat{\gamma} \rho_\vtheta^2(\vz,\vr) + \left(  \frac{\partial \rho_\vtheta(\vz,\vr)}{\partial r_{i,d}}\right)^2\right]}_{\text{aleatoric}} %
	+ \underbrace{\Var_\vtheta\left( \frac{\partial E_\vtheta(\vz,\vr)}{\partial \evr_{i,d}}  \right)}_{\text{epistemic}}.
	\label{eq:force_var_colored_full}
\end{align}


\section{Datasets}\label{app:datasets}
The ANI-1x dataset~\citep{ani1x} consists of Density Functional Theory (DFT) calculations for approximately 5 million diverse molecular conformations with an average of 8 heavy atoms (C, N, O) and an average of 15 total atoms (including H) computed at the \textomega B97x/6-31G(d) level of theory and the structures were generated by pertubing equilibrium configurations and using an active learning procedure to ensure diversity.
The Transition1x dataset~\citep{schreiner2022} contains DFT calculations of energy and forces for 9.6 million molecular conformations with up to 7 heavy atoms (C, N, O) and an average of 14 total atoms (including H), likewise computed at the \textomega B97x/6-31G(d) level of theory and the 
structures were generated by running a nudged elastic band (NEB) algorithm to find transition states between two equilibrium configurations.
Intermediate images of the NEB algorithm were also included in the dataset, with the aim to improve ML potentials around transition states.
The ANI-1x and the Transition1x datasets are split into training, test and validation using the splits from~\citet{schreiner2022}.
For Transition1x the training, validation and test splits contain 9091788, 269636 and 283316 molecules respectively.
For ANI-x the training, validation and test splits contain 4449806, 244331 and 261868 molecules respectively.

\section{Model Hyperparameters and Training}
The models are trained using the Adam optimizer~\citep{kingmaAdamMethodStochastic2017} with an initial learning rate of $10^{-4}$ and a batch size of 64 molecules. The learning rate decays with the factor $0.96^{s/10^5}$ where $s$ is the number of gradient steps.
Simultaneous training of mean and variance networks with negative log-likelihood cost function is an ill-posed problem and requires some tricks to work reliably \cite{skafteReliableTrainingEstimation2019b, seitzer2022betanll}.
When training a model with variance network, we use the following procedure.
The first 2 million gradient update steps we train only the mean function with mean squared error cost function.
For the next 1 million gradient update steps we include training of the variance using the $\beta$-negative-log-likelihood cost function \cite{seitzer2022betanll}.
During this phase the $\beta$ parameter is decreased linearly from 1 (the gradient of the mean error corresponds to squared-error) to 0 (the cost function is identical to negative log-likelihood), which gives a smooth transition from mean squared error to negative log-likelihood and increases the stability of the training.
Finally we train the model for another 3-4 million gradient steps with only the negative log-likelihood cost function.
We use the validation set for model selection using an early stopping procedure and the model selection is always performed using the negative log-likelihood cost function, except for the vanilla model that uses the mean squared error cost function for all training and validation.

Training a single model takes up to 7 days on a single NVIDIA RTX 3090 GPU.
Individual members of the ensemble models are trained in parallel using the same procedure as the single models but with different random seeds for the initialization of the weight matrices and different seeds for the minibatch sampling of the stochastic gradient descent to induce model diversity.

For the PaiNN~\citep{painn} base model we use 3 layers of message passing and a hidden node size of 256. The cutoff distance for creating the message passing graph is 5 Ångström.

\section{Reliability Diagrams and Uncertainty Metrics}\label{app:reliability}
There are various techniques available for assessing the calibration of regression models~\cite{buskGraphNeuralNetwork2023}. The negative log-likelihood (NLL) can be used as a standard metric for evaluating the overall performance of probabilistic models. In that case it measures the (negative) likelihood of observing the hold out test data based on the predicted distribution.
The NLL loss for energy, assuming a normally distributed error, is given for a single instance by the following expression where $x=\{(Z_i,\vec{r_i})\}$ represents the model input and the observed values of energy and forces are denoted by $E^{\text{obs}}$ and $F^{\text{obs}}$, respectively:
\begin{align}
\text{NLL}_E(\theta) &=
-\log p(E^{\text{obs}}|x,\theta) \\
&= 
\frac{1}{2} \Bigg( \frac{\big( E^{\text{obs}}-E(x)\big)^2}{\sigma_E^2(x)}
+ \log \sigma_E^2(x) + \log2\pi \Bigg)
\, .
\label{eq:nll_energy}
\end{align}
The instance-wise energy losses are then averaged over the number of instances.

Analogous to the MSE loss for forces, the NLL loss for forces is evaluated per atom $i$ and component-wise over the spatial dimensions $D$ (recall that the predicted atom-wise force uncertainty $\sigma_{F_i}^2$ is a single scalar applied over all spatial dimensions):
\begin{align}
\text{NLL}_{F_i}(\theta) &=
\sum_{d=1}^D -\log p(F^{\text{obs}}_{i,d}|x,\theta) \\
&= 
\sum_{d=1}^D
\frac{1}{2} \Bigg( \frac{\big( F^{\text{obs}}_{i,d}-F_{i,d}(x)\big)^2}{\sigma_{F_i}^2(x)}
+ \log \sigma_{F_i}^2(x) + \log2\pi \Bigg)
\, .
\label{eq:nll_force}
\end{align}
When using an ensemble of models we use the predicted total mean and total variance to parameterise a normal distribution \cite{busk2021} in order to obtain NLL scores of the test data.

As shown in \eqref{eq:nll_energy} and \eqref{eq:nll_force}, the NLL is dependent on both the predicted mean and uncertainty. However, in some cases, it is more informative to assess the quality of the uncertainty estimates separately.
For instance, a common practice is to visually compare the predicted uncertainties with empirical errors through plotting. Recently \citet{pernot2022} proposed analysis of the z-scores (standard scores). The z-scores are defined as the empirical error divided by the standard deviation of the predictive distribution~\citep{pernot2022}:
\begin{equation}
    z = \frac{y^{\text{obs}}-y(x)}{\sigma(x)} \, .
\end{equation}
To get a single number summarizing the calibration, we can compute the z-score variance (ZV)~\citep{pernot2022}:
A z-score variance (ZV) close to 1 indicates that the predicted uncertainty, on average, corresponds well to the variance of the error, thus indicating good overall calibration. It is often useful to plot the histogram of z-scores, but to summarise the calibration we give the square root of the z-variance (RZV), i.e. the standard deviation of the z-scores.

\begin{figure}[tbp]
	\centering\includegraphics[width=1.\textwidth]{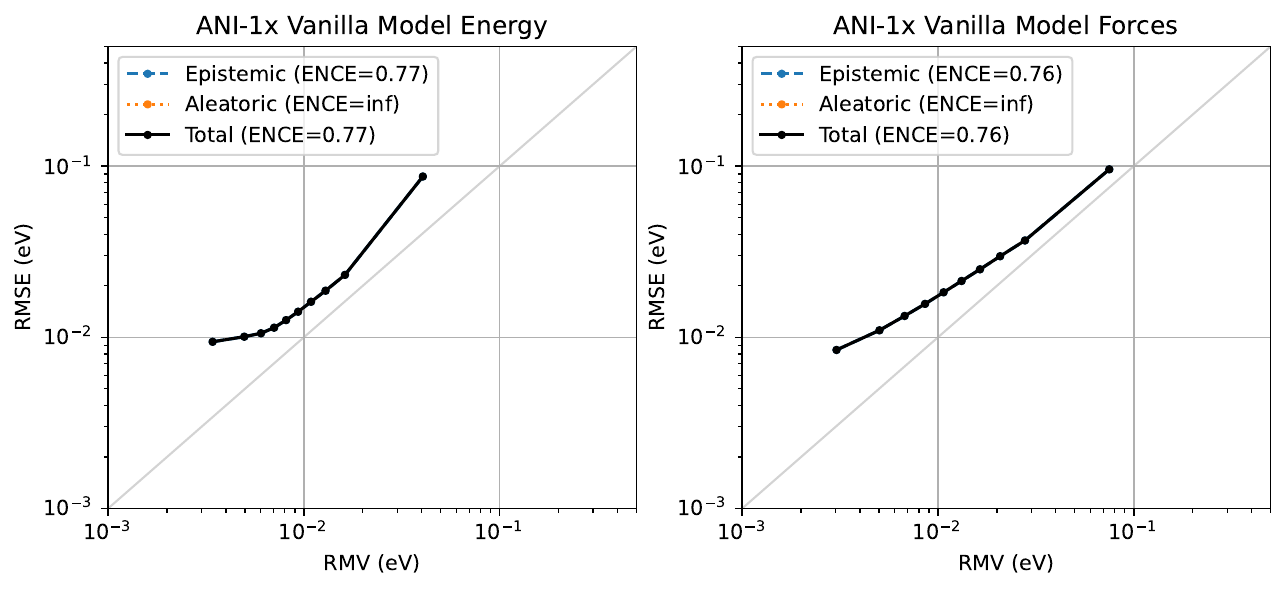}
	\centering\includegraphics[width=1.\textwidth]{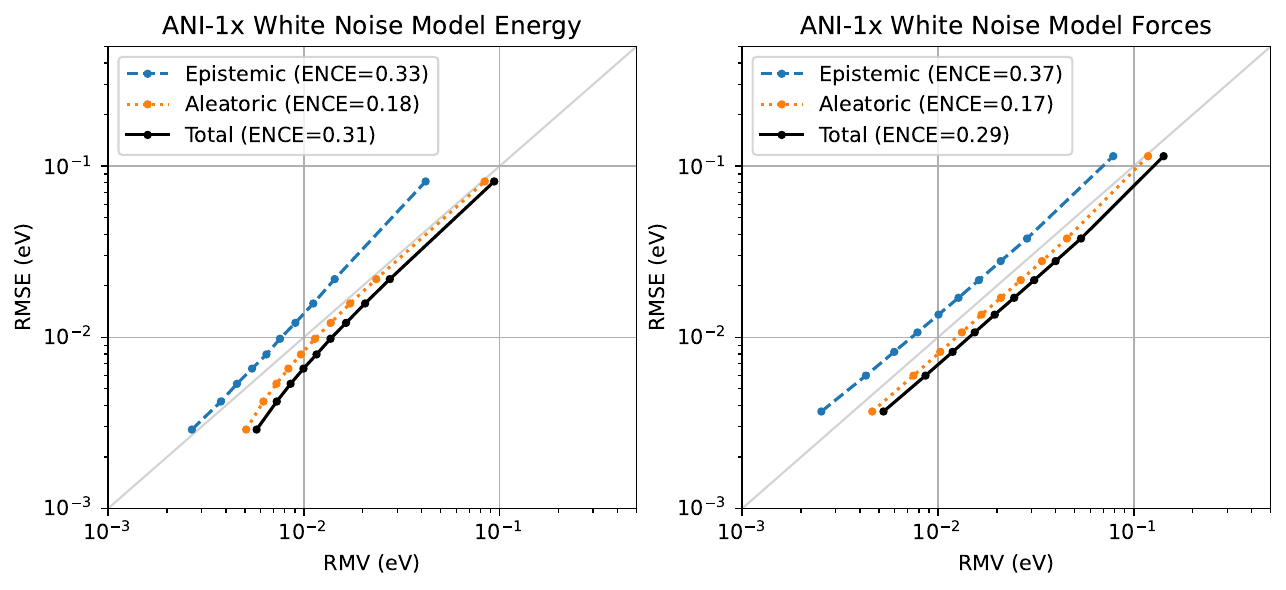}
	\centering\includegraphics[width=1.\textwidth]{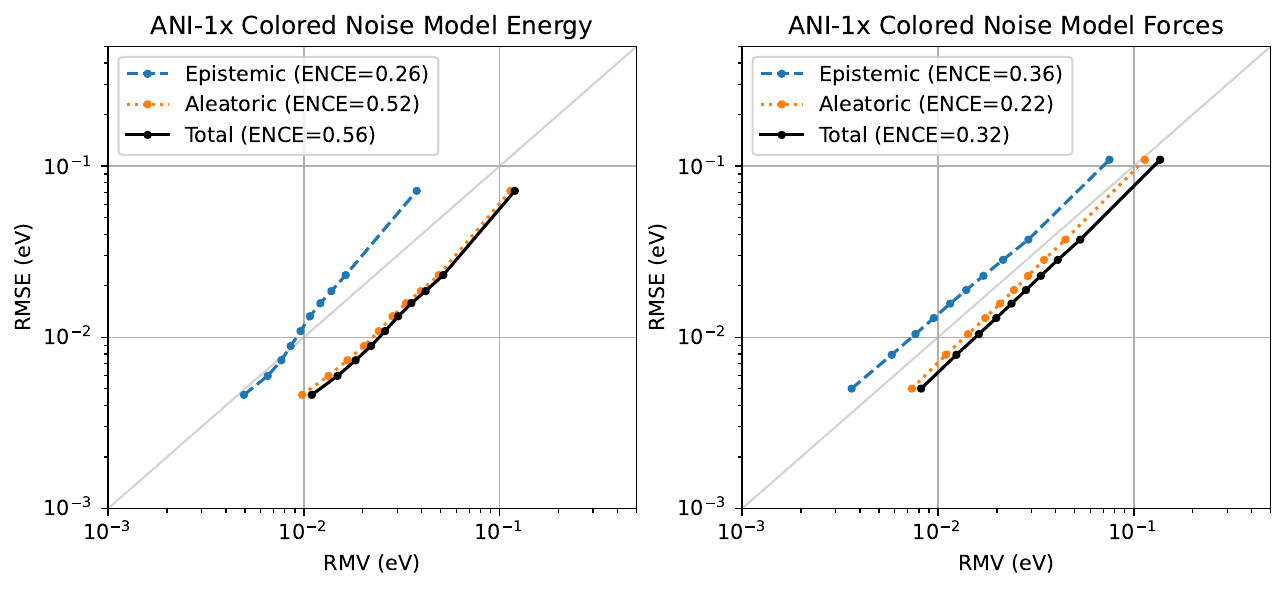}
	\caption{Reliability diagrams for models trained on the ANI-1x dataset.}
	\label{fig:ani-1x}
\end{figure}

\begin{figure}[tbp]
	\centering\includegraphics[width=1.\textwidth]{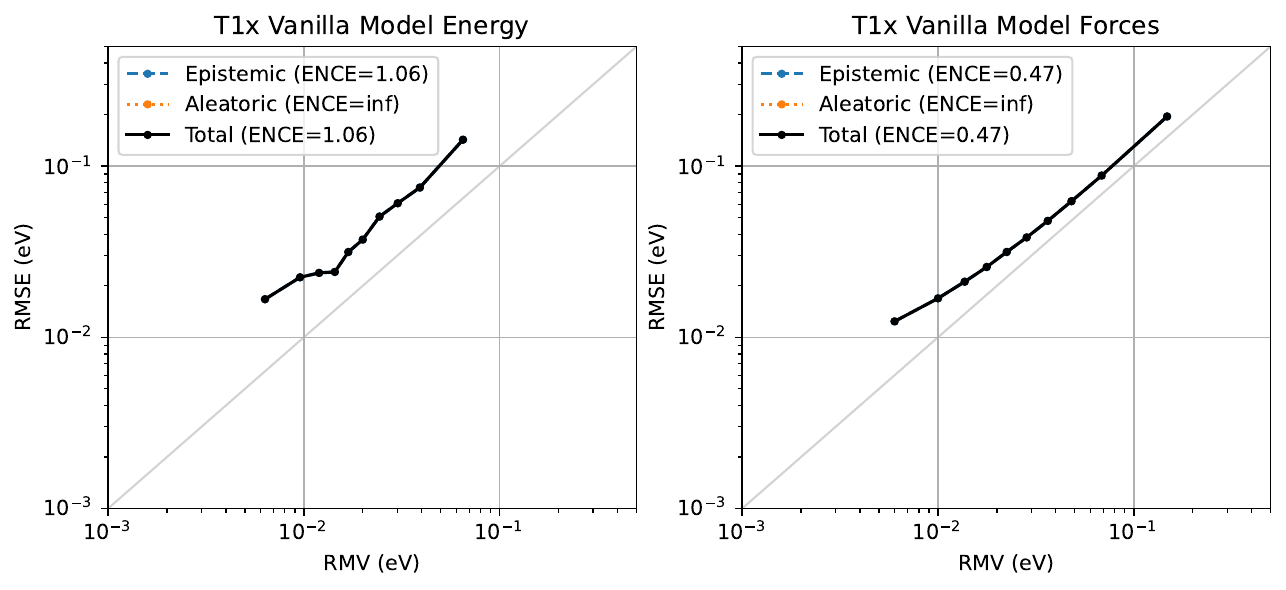}
	\centering\includegraphics[width=1.\textwidth]{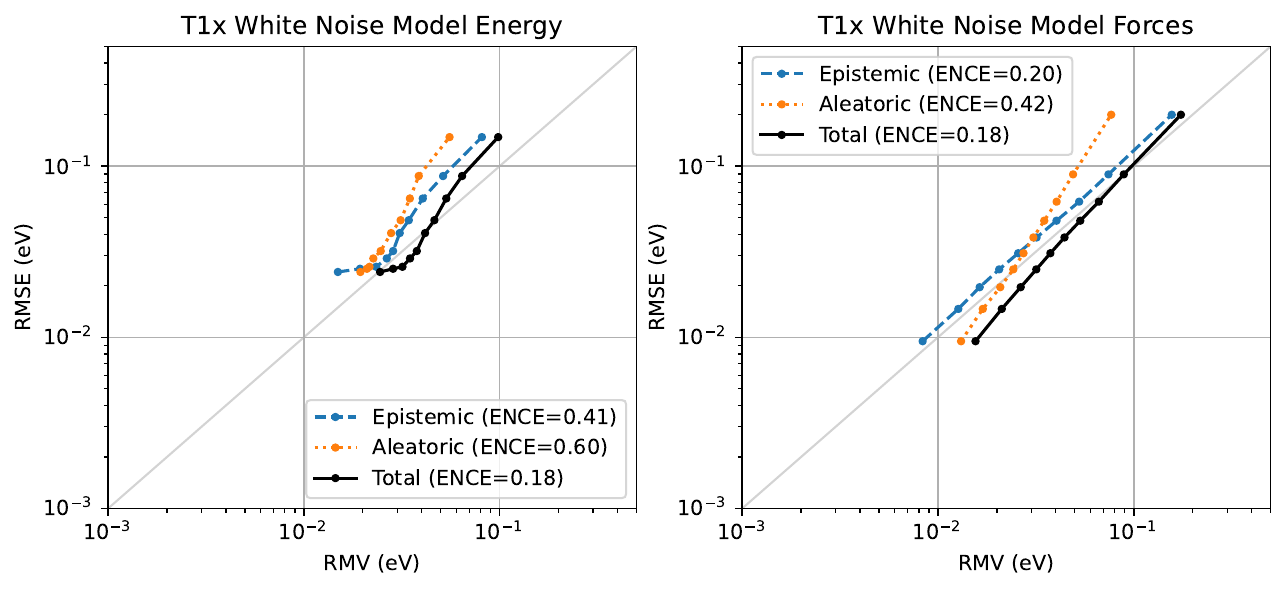}
	\centering\includegraphics[width=1.\textwidth]{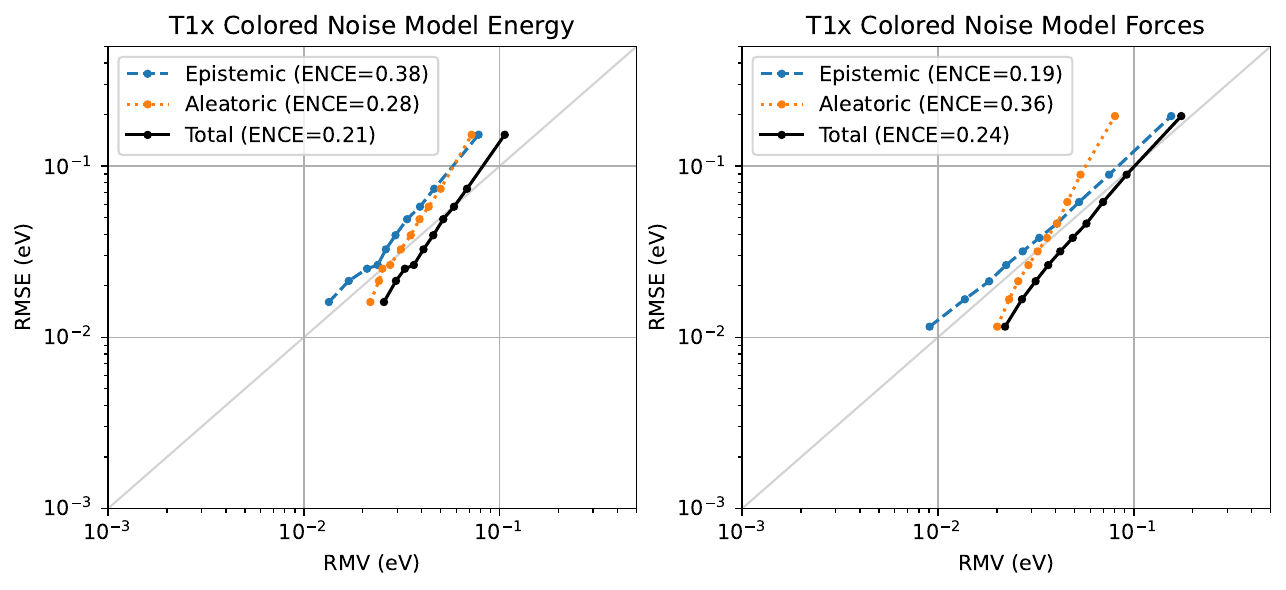}
	\caption{Reliability diagrams for models trained on the Transition1x dataset.}
	\label{fig:t1x}
\end{figure}

Additionally, we can assess how well the uncertainty estimates correspond to the expected error locally by sorting the predictions in equal size bins by increasing uncertainty and plotting the root mean variance (RMV) of the uncertainty versus the empirical root mean squared error (RMSE), also known as an error-calibration plot or reliability diagram. Reliability diagrams of our experiments are shown for ANI-1X in \figurename~\ref{fig:ani-1x} and Transition1x in \figurename~\ref{fig:t1x}
The error-calibration can be summarized by the expected normalized calibration error (ENCE)~\citep{levi2022}, which measures the mean difference between RMV and RMSE normalised by RMV:
\begin{equation}
    \text{ENCE} = \frac{1}{K}\sum_{k=1}^K \frac{|\text{RMV}_k-\text{RMSE}_k|}{\text{RMV}_k} \, ,
\end{equation}
where $k=1,\dots,K$ iterates the bins.

Achieving good average or local calibration is not enough to ensure that individual uncertainty estimates are informative.
If the uncertainty estimates are homoscedastic (lack variation), they may not be very useful.
Therefore, it is generally desirable for uncertainty estimates to be as small as possible while still displaying some variation, which is referred to as sharpness.
To measure sharpness, two metrics can be used: the root mean variance (RMV) of the uncertainty, which should be small and correspond to the RMSE, and the coefficient of variation (CV), which quantifies the ratio of the standard deviation of the uncertainties to the mean uncertainty providing a simple metric for the heteroscedasticity of the predicted uncertainty. The equation for CV is as follows:
\begin{equation}
    \text{CV} = \frac{\sqrt{N^{-1} \sum_{n=1}^N (\sigma(x_n) - \overline\sigma)^2}}{\overline\sigma} \, ,
\end{equation}
where $n=1,\dots,N$ in this case iterates the test dataset, $\sigma(x_n)$ is the predicted standard deviation (uncertainty) of instance $n$ and
$\overline\sigma = N^{-1} \sum_{n=1}^N \sigma(x_n)$
is the mean predicted standard deviation.


\end{document}